\documentclass{article}
\usepackage{arxiv}

\usepackage[utf8]{inputenc} 
\usepackage[T1]{fontenc}    
\usepackage{hyperref}       
\usepackage{url}            
\usepackage{booktabs}       
\usepackage{amsfonts}       
\usepackage{nicefrac}       
\usepackage{microtype}      
\usepackage{graphicx}
\usepackage{amsmath}
\usepackage{subfigure}
\usepackage{graphicx}  
\title{Learning to Answer Ambiguous Questions with knowledge graph}

\author{
Yikai Zhu\textsuperscript{\rm 1} , Jianhao Shen\textsuperscript{\rm 1}, Ming Zhang\textsuperscript{\rm 1} \\
\textsuperscript{\rm 1}School of EECS, Peking University\\ 
zhuyikai.zyk@gmail.com, jhshen@pku.edu.cn, mzhang@net.pku.edu.cn
}
\begin{document}
\maketitle

\begin{abstract}
In the task of factoid question answering over knowledge base, many questions have more than one plausible interpretation. Previous works on SimpleQuestions assume only one interpretation as the ground truth for each question, so they lack the ability to answer ambiguous questions correctly. In this paper, we present a new way to utilize the dataset that takes into account the existence of ambiguous questions. Then we introduce a simple and effective model which combines local knowledge subgraph with attention mechanism. Our experimental results show that our approach achieves outstanding performance in this task.
\end{abstract}
\section{Introduction}
Knowledge bases (KBs), such as Freebase~\cite{bollacker2008freebase}, DBPedia~\cite{lehmann2015dbpedia} and YAGO~\cite{suchanek2007yago}, comprise vast amounts of facts about the real world. KBs are usually not easily accessible for ordinary users as they need to be familiar with query languages (e.g. SPARQL~\cite{prud2006sparql}) as well as the structure and relations in KBs. Factoid question answering over knowledge bases (KBQA), which allows users to query KBs with natural language, provides a more easy-to-use access to KBs and has attracted much attention recently~\cite{berant2013semantic,yih2015semantic,xu2016question,golub2016character}. WebQuestions~\cite{berant2013semantic} and SimpleQuestions~\cite{bordes2015large} are the most commonly used datasets for KBQA, and in this work we mainly focus on SimpleQuestions since it has a much larger scale than WebQuestions and single-relation questions are the most common type of questions observed in community QA sites~\cite{fader2013paraphrase}

One drawback of using natural language instead of structured query language is its ambiguity --- many questions have more than one plausible interpretation. The KBQA system usually performs two key tasks: (1) entity detection, which links n-grams in questions to KB entities, and (2) relation prediction, which identifies the KB relations that a question refers to. Ambiguity comes from both entity detection, where multiple nodes can share the same name, and relation prediction, where one question pattern may have multiple meanings. Ambiguous questions are very common in KBQA: we found that there are at least 33.9\% questions in SimpleQuestions and 37.6\% questions in WebQuestions are ambiguous. 

A decent way to answer these ambiguous questions is to first check the user's intention by asking a question like "Which one do you mean, A, B or C?" and give the final answer based on the user's response. This requires KBQA systems being able to find all plausible interpretations for one question. However, each question in SimpleQuestions is labeled with only one interpretation as the ground truth, and previous works based on that are evaluated by their accuracy of predicting this ground truth, so the problem of ambiguity is ignored. In this work, we treat SimpleQuestions in a new way by automatically labeling each question with its plausible interpretations based on original labels, and use F1 score as the evaluation metric. This formulation benefits further research on ambiguous question answering.

Then we introduce a simple and effective model which achieves outstanding performance in this task. The entity detection part of our model follows ~\cite{petrochuk2018simplequestions}.For each entity candidate, we use GRU to get the embedding of its one-hop subgraph, and feed the embedding of subgraph as well as the question into a relation predictor. The predictor uses attention mechanism to focus on the most relevant part of the question with the subgraph relations, and predicts the possibility for each relation in the subgraph. Our approach achieves 84.9\% F1 score for predicting all subject-relation pairs.

Our main contribution includes: (1) we present a new way to use and evaluate on SimpleQuestions that takes into account the ability of answering ambiguous questions. (2) we introduce an effective model which achieves outstanding performance in ambiguous question answering.

The rest of the paper is structured as follows. Section II introduces related work. The motivation of this study and some notation are given in Section III. The details of the proposed algorithm are provided in Section IV. The performance of our model is demonstrated in Section V. Section VI concludes this study.

\section{Related works}
There are two main lines of solutions to KBQA: semantic parsing based (SP-based) approaches and information retrieval based (IR-based) approaches.

SP-based approaches\cite{berant2013semantic,yih2014semantic,yih2015semantic} aim to learn semantic parsers which parse natural language questions into structured queries (e.g. logical forms) and then query knowledge base to look up answers. These approaches are difficult to train at scale because of the complexity of their inference.

IR-based approaches\cite{petrochuk2018simplequestions,yu2017improved} try to extract useful information and features from both the question and KB knowledge, and obtain target answers by matching the question and answers using these features. 

Recently, as deep neural networks have been applied to many NLP tasks, NN-based approaches has attracted more and more attention in KBQA\cite{bordes2014question,bordes2015large,golub2016character,lukovnikov2017neural}. NN-based approaches share the same idea with IR-based approaches. The main difference is that they use neural networks to encode questions, relations and answers into low-dimensional continuous vectors. The answer which has the highest matching score with the question will be chosen. Our model also falls in this framework.

\section{Ambiguity}
In this section, we first introduce some basic notation. Then we will define what is ambiguity in the knowledge graph. 

\subsection{Notation}
In a knowledge graph $\mathcal{KB}$, we use $\{a_1, ..., a_n\}$ to define all the aliases of a entity. For example, "malcolm x" is one alias for the entity "/m/04vthmn" in FB2M. For an entity $e$, we define $\{ (e, r, t) | (e , r, t) \in \mathcal{KB} \}$ as the one-hop subgraph from entity $e$. We use $(e, r, *) \in \mathcal{KB}$ to denote that $\exists e', (e, r, e') \in \mathcal{KB}$.

Given a example question $q$, we want to find a path $(e_0, r_1, e_1, r_2, e_2,..., r_n, e_n)$ in the knowledge graph $\mathcal{KB}$, such that $\forall i \in \{1, ..., n\}, (e_{i-1}, r_i, e_i) \in \mathcal{KB}$, $q$ can map to the name of $e_0$ and $(r_1, ..., r_n)$ and $e_n$ is the answer to the question. A dataset $\mathcal{D}$ can be denoted as $\mathcal{D} = \{ (q, (e_0, r_1, ..., r_n, e_n)) \} $.We define the formatted question $\hat{q}$ of $q$ with the subject alias $a$ replaced by $\langle e\rangle$. For example, for the question "who wrote malcolm x", "who wrote $\langle e\rangle$" is its formatted question. If $n=1$, we define such question is a single-relation question. Our research is mainly on single-relation questions. 

\subsection{Definition}
Generally, there are two kinds of ambiguities, entity-level ambiguity and relation-level ambiguity.

\subsubsection{entity-level ambiguity} Entity-level ambiguity is that a mention in the question may correspond to multiple distinct entities in the underlying knowledge graph. For example, in the question "who is the author of malcolm x", the entity "malcolm x" may map to a book, a movie or music. In Table \ref{description}, we can see different kinds of malcolm x and the number of nodes with corresponding types in FB2M.

\subsubsection{relation-level ambiguity} Relation-level ambiguity is that in a dataset $\mathcal{D}$, for one formatted $\hat{q}$, there exists multiple relation $r$, such that $(\hat{q}, r) \in \mathcal{D}$, which means one formatted question may map to different relations. For example, the formatted question "who wrote $ \langle e \rangle $" may map to \texttt{book/book\_edition/author\_editor} or \texttt{music/composition/composer}. In Table \ref{relation} we can find that a simple formatted question as "who is the author of $\langle e \rangle$" can map to multiple relations.

If for a question $q$, $\exists e_1, e_2, r_1, r_2$, such that $ (e_1, r_1, *) \in \mathcal{KB}, (e_2, r_2, *) \in \mathcal{KB}, (e_1, r_1) \neq (e_2, r_2)$, $q$ can map to $e_1$ and $e_2$ and $\hat{q}$ can map to $r_1$ and $r_2$, we can say that this question is ambiguous since we can find multiple answers for this question. We find that 33.9\% of examples in the SimpleQuestions dataset and 33.7\% of examples in the WebQuestions dataset are ambiguous. In fact, in reality, ambiguous questions is pretty common and we are supposed to find a way to answer them.

\begin{table}[t!]
\begin{center}
\begin{tabular}{|l|r|}
\hline \bf Description & \bf count \\ \hline
malcolm x(book) & 22    \\
malcolm x(music record) & 14  \\
malcolm x(film) & 2 \\
malcolm x(song)  & 1 \\
malcolm x(other type) &  7 \\ 
\hline
\end{tabular}
\end{center}
\caption{\label{description} different descriptions of malcolm x and the number of nodes with corresponding description in the FB2M}
\end{table}

\begin{table}[t!]
\begin{center}
\begin{tabular}{|l|r|}
\hline \bf reasonable relation & \bf count \\ \hline
common/topic/notable\_types & 51 \\
book/book\_edition/author\_editor & 19  \\
music/recording/artist  & 11 \\
book/written\_work/author & 3 \\
book/author/works\_written & 3 \\ 
music/album/artist & 2\\
music/composition/composer  & 1 \\
\hline
\end{tabular}
\end{center}
\caption{\label{relation} reasonable relations to the question "who is the author of malcolm x" and the number of answers each relation can lead to in the SimpleQuestions dataset}
\end{table}

\section{Model}
The model can be divided into two parts. The first part maps question to entities and the second part maps formatted question to relation. The first part gives every word in the question a tag of 0 or 1. These words with tag 1 are regarded as entity alias and these words with tag 0 are regarded as formatted question. There have been lots of efforts devoted to named-entity recognition part and the first part is not the bottleneck of KBQA. We simply use the BiGRU-CRF model provided by \cite{petrochuk2018simplequestions} as our first part. This model can predict the accurate alias $a$ for 95.5\% of the questions in the test set, which is pretty good. Then for each entity candidate, its one-hop relations and the formatted question are fed into the second part to predict relation. We propose Knowledge Subgraph with Attention BiGRU (KSA-BiGRU) model as the second part. In this section we describe our model in details.
\begin{figure*}[!t]
  \includegraphics[width=\linewidth]{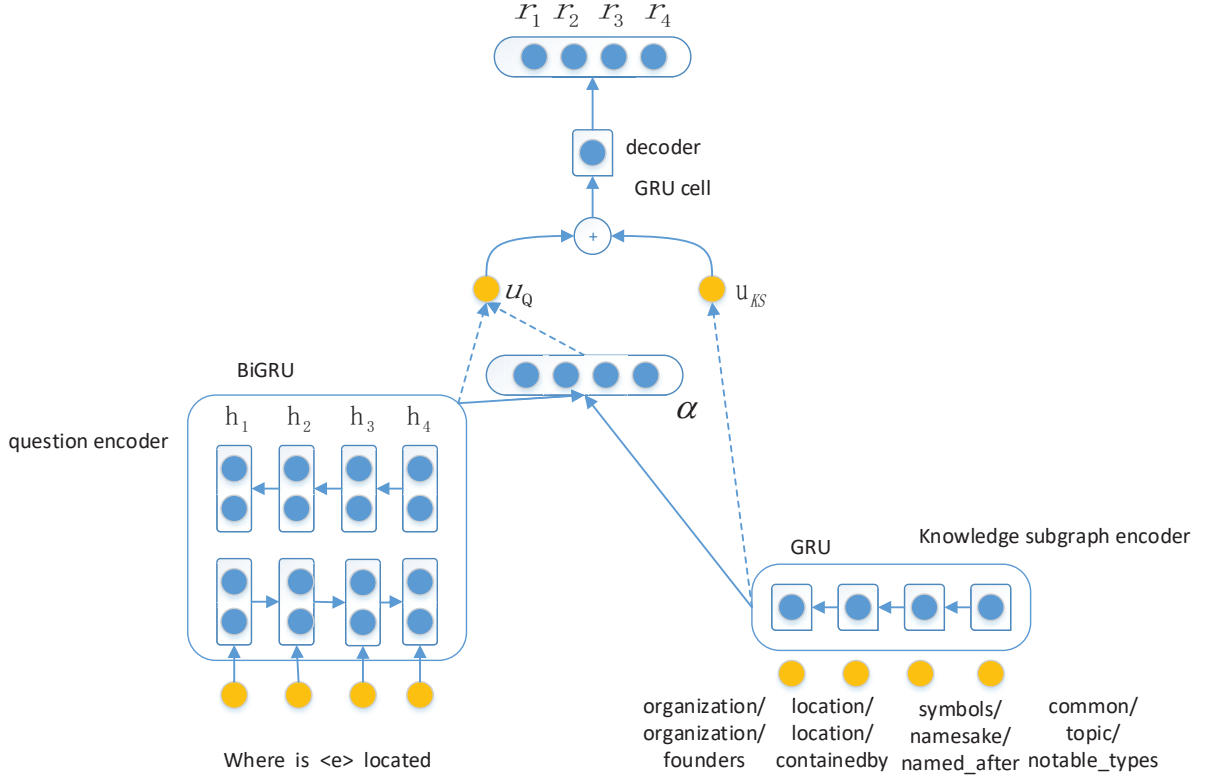}
  \caption{the architecture of KSA-BiGRU}
  \label{fig:architecture}
\end{figure*}

\subsection{Data \& evaluation metric}
\cite{petrochuk2018simplequestions} treat ambiguous questions as unanswerable and give a upper bound of performance at 83.4\%. Different from their work, we consider all the reasonable pairs as the candidates, and view each question as a multi-label data.
We want our model to find all the reasonable semantic interpretations among all the potential interpretations. One obvious advantage is our negative sampling way.In traditional methods of answering single-relation questions, for $(q, (s, r, o))$, first they use entity detector and get $\hat{q}$ and $s$. After that, they consider all the relations in the subgraph of $s$ except $r$ as negative. This will cause problem because they don't consider other data in the dataset and there might be other reasonable relations in the subgraph as well. For example, for question "who wrote malcolm x", both the relations \texttt{music/recording/artist} and \texttt{music/composition/composer} are in the subgraph of malcolm x(music record). In fact, both of these two relations might be reasonable.We should include neither of them in the negative sampling. We know that both ("who wrote $\langle e \rangle$", \texttt{music/recording/artist}) and ("who wrote $\langle e \rangle$", \texttt{music/composition/composer}) appear in the dataset. Traditional ways simply ignore the rest of the dataset while dealing with one data.

To achieve this goal, 
instead of predicting $p(s, r | q)$, our model learns a distribution $ p(y|q, s,r) $, where y takes value in $\{0, 1\}$ and $p(y = 1|q, s, r)$ represents the probability of $(s, r)$ being a reasonable interpretation of q. Then the final prediction of our model is $\{ (s,r) | p(1|q,s,r)>0.5 , s \in \mathcal{S} \land (s,r) \in \mathcal{KG}\}$, where $\mathcal{S}$ is the set of all entities share the same alias. Precision, recall and F1 are computed based on the output for each individual question and the average F1 score is reported as the main evaluation metric.

As mentioned before, our model learns a conditional distribution $ p(y|q, s,r) $. It needs to be calculated over all pairs (s, r) for a given question, which is problematic. Notice that if $s$ is not in the reasonable subjects set of $q$ or $r$ is not connected to $s$ in the KB, then $p(y=1|q,s,r)=0$. Therefore we only need to focus on entities that might be aligned to the question and their one-hop relations.

\subsection{Model Details}

Given a formatted question $\hat{q}$ and a subject $s$, we use a encoder-decoder structure to compute $p(y|q,s,r)$ for every relation $r$. The model uses the encoding of all the relations around the subject, which form a subgraph of the knowledge base, to represent the information from the subject. In addition, we use a BiGRU with attention to encode the question. Figure \ref{fig:architecture} shows the architecture of our model.

{\bf Subgraph Encoder} 
For the subject $s$, let $R(s) = (r_1, r_2, ..., r_n)$ denote all the relations which have connections with subject $s$ in the knowledge base.  We use an embedding layer to get the vector representation for every relation and then we take them as the input sequence to GRU. The initial state of the network is a zero vector. We denote the final state of the GRU network by $u_{KS}$, which we refer to as the encoding of the knowledge subgraph. Although the order of the relations in $R(s)$ should not matter, we still feed them to the GRU network sequentially because the GRU network has the ability to process variable-length input and automatically keep relevant features. In fact, we shuffle the order of the relations in the experiments and find that the result doesn't change a lot.

{\bf Question Encoder} For a question $q$, let $Q=(v_1, ..., v_m)$ be all the words in the formatted question $\hat{q}$. We use an embedding layer to get the word embeddings of every word in Q. These embeddings are fed into a  bidirectional GRU network to get hidden representations $ [h_1, ..., h_m] $ (each vector $h_i$ is the concatenation between forward/backward hidden state at time $i$). 

In addition, since the formatted question is the key to choose the answers, we want our model to focus on essential words in the question. For example, for the formatted question \texttt{"who is the author of <e>?"}, we think our model should focus more on the word ``author" than the word ``is". Therefore, we use an attention layer to summarize the query and context into single feature vectors. Here we regard $[h_1, ..., h_m]$ as the context and $u_{KS}$ as the query. The formatted question representation $p$ in calculated as :
\begin{align}
    p &= \sum_{j=1}^m \alpha_{j}h_{j} \\
    \alpha_j &= \frac{exp(w_j)}{\sum_{k=1}^{m}exp(w_k)} \\
    w_j &= v^{T} f(W^T[h_j;u_{KS}] + b)
\end{align}
where $\alpha_j$ denotes the attention weight of the j-th word in the question, f is the activation function. Let the dimension of  $h_i$ be $d_{seq}$ and the dimension of $u_{SK}$ be $d_{KS}$. $W$ and $v$ are the parameters to be learned with $ W \in R^{c\times(m+n)}, v \in R^{1 \times c} $, where c is the size of the hidden state in the attention layer, which is a hyperparameter.

{\bf Decoder and Answer Selector} We use a single GRU cell as the decoder for relation decoding. The output of the encoder is fed into the GRU cell as the initial hidden state. We use the embedding of a special symbol \texttt{<\_start>} as the input of the GRU. The output of the GRU is fed into a fully connected layer and then passed to a sigmoid function to get the probability of each relation to be a plausible relation, which we denote as $p(y=1|q,s,r)$. We can set a threshold $\lambda$ and choose relations whose probabilities are larger than  $\lambda$ as the answers. We choose $\lambda=0.5$ in this work.

Under the maximum-likelihood principle, the loss function of KSA-BiGRU is :
\begin{equation}\begin{aligned}
    Loss(p)=&-\sum_{(q, s, r^{+})\in D^{+}}log(p(y=1|q,s,r^{+}))\\
    & -\sum_{(q,s,r^{-}) \in D^{-}}log(p(y=0|q,s,r^{-}))
\end{aligned}
\end{equation}

where $D^{+}$ is the set of all positive examples and $D^{-}$ the set of negative examples. For each positive example $(q,s,r^{+})$, we sample 5 negative examples $(q,s,r^{-})$.

\section{Experiments}
\label{sec:length}
There are mainly two large datasets on KBQA, SimpleQuestions and WebQuestions. Since SimpleQuestions only contain single-relation questions and WebQuestions contain question with multi-hop relation questions. Evaluation was carried out on SimpleQuestions. We carry out our experiments on one dataset since there are not other datasets in KBQA and a lot of previous works also use SimpleQuestions as the only dataset. SimpleQuestions(SQ) consists of 108,442 questions written by human English-speaking annotators. The dataset is partitioned randomly into 75,910 training questions, 10,845 validation questions and 21,687 test questions. The questions in SQ are all single-relation questions.

A subset of Freebase, FB2M \cite{bordes2015large}, is also used in our study as the knowledge base. It includes 2,150,604 entities, 6701 relations and 14,180,937 triples. All the answer triples in SQ can be found in FB2M.

\begin{table*}[t!]
\begin{center}
\begin{tabular}{|l|rrr|}
\hline \bf Method & \bf Rec. & \bf Prec. & \bf F1 \\ \hline
random guess & 49.2 & 19.4 & 24.0 \\
BiLSTM\cite{petrochuk2018simplequestions} & 81.2 & 82.6 & 80.9  \\
\hline
BiGRU (ours) & 82.3 & 81.0 & 80.7 \\
KS-BiGRU (ours) & 84.9 & 82.4 & 82.7 \\
KSA-BiGRU (ours)& {\bf 86.7} & {\bf 84.8} & {\bf 84.9} \\
\hline
\end{tabular}
\end{center}
\caption{\label{table:F1} The results of our approach compared to existing work. The numbers of other systems are derived from the evaluation script }
\end{table*}

\begin{table*}[t!]
\begin{center}
\begin{tabular}{|l|r|}
\hline \bf Method & \bf acc \\ \hline
random guess & 4.9 \\
Memory NN\cite{bordes2015large} & 71.6 \\
AR-SMCNN\cite{qu2018question} & 72.4 \\
BiLSTM\cite{petrochuk2018simplequestions} & {\bf 78.1}  \\
\hline
KSA-BiGRU (ours) & 73.1 \\
\hline
\end{tabular}
\end{center}
\caption{\label{table:acc}  compared to existing work in traditional ways. The numbers of other systems are either from the original papers or derived from the evaluation script, when the output is available}
\end{table*}
\subsection{Implementation}

In this work, our entity alignment model uses the recent LSTM-CRF model  with conditional log likelihood loss objective\cite{petrochuk2018simplequestions}. The entity alignment model is trained separately. \footnote{For more details, source code is available at https://github.com/PetrochukM/Simple-QA-EMNLP-2018}

During training, all the word embeddings are randomly initialized with 500 dimensions. We have tried pretrained word embeddings but it gives a sightly worse result. The embeddings of relations are initialized using TransE\footnote{For more details, source code is available at https://github.com/DeepGraphLearning} with 300 dimensions, which is trained separately. The size of hidden states for each of the two GRU networks is set to 300. Adam, initialized with a learning rate of 0.001, is employed to optimize the model. Dropout is used to regularize the two-layer BiGRU in our experiment and the drop rate is set to 0.1. The size of hidden state $c$ in the attention layer is set to 650.

\subsection{Overall results}
Experiments are performed to evaluate the proposed KSA-BiGRU model in our ambiguous question answering tasks, relative to other existing models and variations of KSA-BiGRU. Results are shown in table \ref{table:F1}. Our KSA-BiGRU outperforms the state-of-the-art model\cite{petrochuk2018simplequestions} and achieves 84.9 \% F1.   \\
\begin{figure}
  \includegraphics[width=\linewidth]{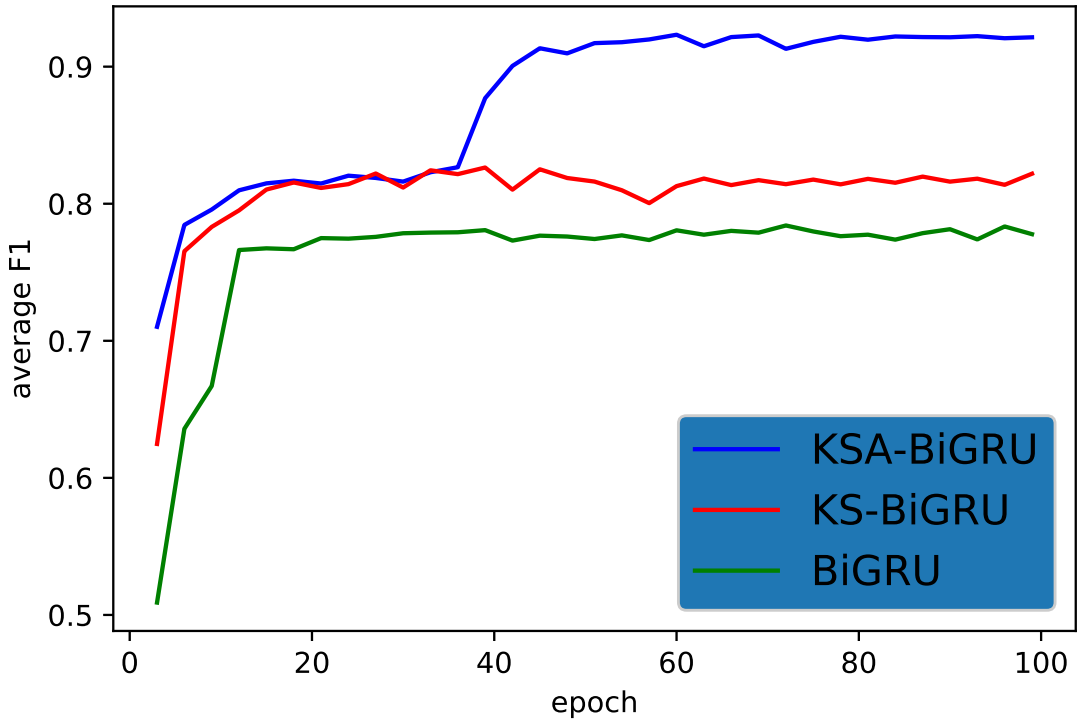}
  \caption{the performance of KSA-BiGRU and its variations}
  \label{fig:F1}
\end{figure}
{\bf Analysis of KSA-BiGRU} Figure \ref{fig:F1} plots the performances of KSA-BiGRU and its variations. KS-BiGRU is the KSA-BiGRU model without attention mechanism. It uses the concatenation of $u_{Q}$ and $u_{KB}$  as output of the encoder and the decoder is the same as KSA-BiGRU. BiGRU is a reduction of KSA-BiGRU by completely removing the part of knowledge subgraph, simply using $u_{Q}$ as the output. Comparing the performance of KS-BiGRU with BiGRU, we can find that including the information of knowledge subgraph appears to give a much significant boost of the performance. Adding attention mechanism boosts the performance further, reaching F1 of 84.9\% on test set. In addition, the performance of KSA-BiGRU over epochs can also be shown in Figure \ref{fig:F1}. It can be seen that the performance increases with epochs until about 45 epochs. This demonstrates the effectiveness of our model. 

{\bf Evaluate in traditional ways} In traditional ways, each question q is annotated with one single answer pair $(\hat{s}, \hat{r})$ as the answer. We denote the set of all the plausible answers of q as $ {SR}(q) $. KSA-BiGRU is designed and trained to find the $SR(q)$ while it can also be used to predict $(\hat{s}, \hat{r})$. We can simply choose the entity and relation pair which has the highest prediction probability as the answer. Our model achieve 73.1\% accuracy on SimpleQuestions test set, while the state-of-the-art BiLSTM achieves an accuracy of 78.1\%.
For unambiguous questions, our model can predict predict answers for 78\% of them, while for ambiguous questions, our model will predict another reasonable answer in many cases. In 84.65 \% of the cases, KSA-BiGRU can successfully predict one $(\hat{s}, \hat{r})$ pair among all reasonable semantic interpretations. 

\subsection{Effectiveness of attention}
\begin{figure*}
\subfigure{
\label{Fig:atten1}
\includegraphics[width=\linewidth]{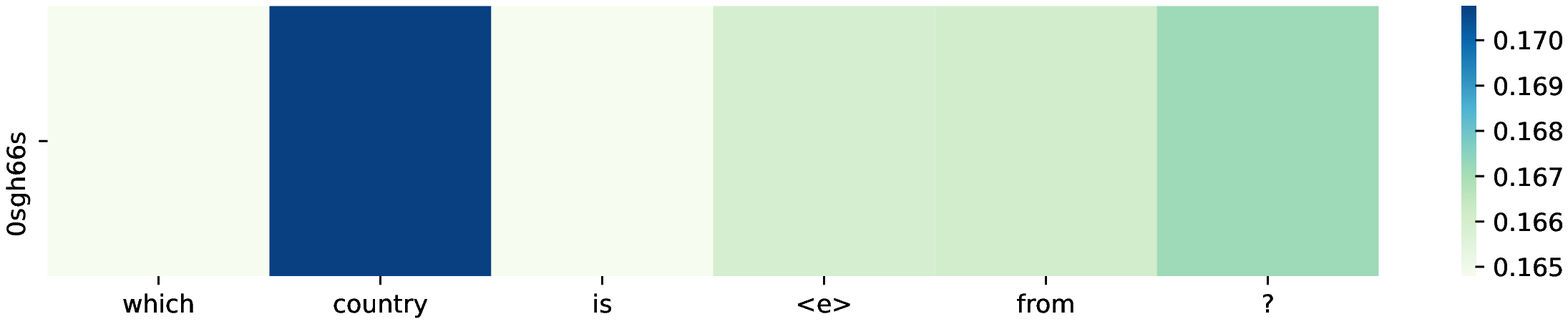}
}
\subfigure{
\label{Fig:atten2}
\includegraphics[width=\linewidth]{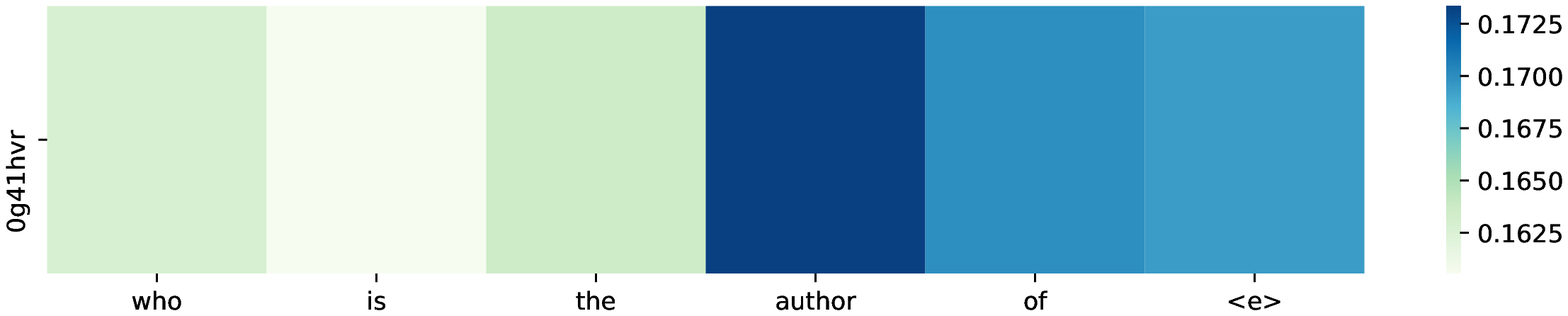}
}
 \caption{The visualized attention map. Depth of the color represents the attention weights,and the darker the higher}
 \label{fig:attention}
\end{figure*}

According to the results in section 5.2,  the attention mechanism plays an essential role in achieving a better performance. To obtain a deeper understanding of its effectiveness, we visualize the attention distribution of the question words in two examples, which is displayed in Figure \ref{fig:attention}.

We can find out that the essential words is put more attention than other words. In the first example ,the attention mechanism helps the encoder pay more attention to the word ``country", which is obviously the key word in the example.It shows that KSA-BiGRU learns a reasonable weight distribution of the question words. 

\subsection{Further Qualitative Analysis}
We also analyze the performance with specific cases. Specially, we analyze the errors to point toward directions for future work.  Before analyzing, we find that the entity alignment model we use failed to detect the subject of 4.5\% of the questions. We remove these data first and analyze the performance of our model on the rest 20710 cases.

For 76.73\%(15891) of all the questions,  the subject-relation pairs which the KSA-BiGRU predict are exactly the same as the plausible subject-relations pairs. For example,  for the question "what country is lamb from?", the standard answer is "film/film/country" and the output of KSA-BiGRU will be \texttt{\{['0g41hvr'(lamb, film),'film/film/country', '07ssc'(britain)],  ['0t\_bhxw'(lisa cashman\footnote{her nickname is lamb},people),'people/person/nationality', '09c7w0'(u.s.a .)],  ['0n91fn2'(tv series),  'tv/tv\_program/country\_of\_origin', '03\_3d'(japan)], ......\}}. All of these answers provided by KSA-BiGRU are reasonable since all these relations are the standard answer of one question with format of \texttt{"what country is <e> from?"}. 

The remaining 23.27\%(4819) error comes from a number of sources. We do an empirical error analysis on a sample of 50 negative examples and identify error cases among them. 

\begin{itemize}
\item {\bf Overconfidence} In 15 of 50 questions, the model chooses one more relation as the plausible relation.  We find the relation chosen by our model may be plausible! For example, the standard answer for the question "what song was on the recording indiana" is \texttt{"music/recording/song"}, while our model gives us one more relation \texttt{"music/release\_track/recording"}. Both of them can lead to the answer \texttt{"indiana"}.  

\item {\bf Ignoration}  In 33 of questions, the model ignores one or two reasonable answers. In 14 questions the model thinks all the candidate relations are negative relations although there is one answer.  Part of the reasons are some formatted questions are so rare that they are seen in the training data less than 8 times.

\item {\bf Noise}  2 of 50 questions just make no sense . For example, the question "how does engelbert zaschka identify" apparently lack a word after "identify". 
\end{itemize}

\section{Conclusions}
In this paper, we use and evaluate on SimpleQuestions in a new way that takes into account the ability of answering ambiguous questions. Then we propose a simple and effective model which leverage both knowledge subgraph embedding and attention mechanism, and achieves outstanding performance in this task. We hope our work would help future research on answering ambiguous questions and building a better KBQA system. 

\bibliography{aaai.bib}
\bibliographystyle{unsrt}
\end{document}